\documentclass[conference]{IEEEtran}
\IEEEoverridecommandlockouts
\usepackage{hyperref}
\usepackage{cite}
\usepackage{amsmath,amssymb,amsfonts}
\usepackage{algorithmic}
\usepackage{graphicx}
\usepackage{multirow}
\usepackage{textcomp}
\usepackage{xcolor}
\def\BibTeX{{\rm B\kern-.05em{\sc i\kern-.025em b}\kern-.08em
    T\kern-.1667em\lower.7ex\hbox{E}\kern-.125emX}}

\begin{document}
\title{Deep object detection for waterbird monitoring using aerial imagery}

\author{
    \IEEEauthorblockN{Krish Kabra\IEEEauthorrefmark{1}\textsuperscript{,}\IEEEauthorrefmark{2}\textsuperscript{,}\IEEEauthorrefmark{3},
    Alexander Xiong\IEEEauthorrefmark{2}\textsuperscript{,}\IEEEauthorrefmark{3}, 
    Wenbin Li\IEEEauthorrefmark{2}\textsuperscript{,}\IEEEauthorrefmark{3}, 
    Minxuan Luo\IEEEauthorrefmark{3}, 
    William Lu\IEEEauthorrefmark{3}, \\
    Tianjiao Yu\IEEEauthorrefmark{3}, 
    Jiahui Yu\IEEEauthorrefmark{3}, 
    Dhananjay Singh\IEEEauthorrefmark{3}, 
    Raul Garcia\IEEEauthorrefmark{3}, 
    Maojie Tang\IEEEauthorrefmark{3}, \\
    Hank Arnold\IEEEauthorrefmark{4}, 
    Anna Vallery\IEEEauthorrefmark{4}, 
    Richard Gibbons\IEEEauthorrefmark{5}, 
    Arko Barman\IEEEauthorrefmark{4}
    }
    
    \IEEEauthorblockA{\IEEEauthorrefmark{3}Rice University, Houston, TX 77005, USA}
    \IEEEauthorblockA{\IEEEauthorrefmark{4}Houston Audubon Society, Houston, TX 77079, USA}
    \IEEEauthorblockA{\IEEEauthorrefmark{5}American Bird Conservancy, The Plains, VA 20198, USA}
    
    \thanks{\IEEEauthorrefmark{2}Denotes equal contribution. \IEEEauthorrefmark{1}Corresponding author: kk80@rice.edu}

}

\maketitle

\begin{abstract}
Monitoring of colonial waterbird nesting islands is essential to tracking waterbird population trends, which are used for evaluating ecosystem health and informing conservation management decisions. Recently, unmanned aerial vehicles, or drones, have emerged as a viable technology to precisely monitor waterbird colonies. However, manually counting waterbirds from hundreds, or potentially thousands, of aerial images is both difficult and time-consuming. In this work, we present a deep learning pipeline that can be used to precisely detect, count, and monitor waterbirds using aerial imagery collected by a commercial drone. By utilizing convolutional neural network-based object detectors, we show that we can detect 16 classes of waterbird species that are commonly found in colonial nesting islands along the Texas coast. Our experiments using Faster R-CNN and RetinaNet object detectors give mean interpolated average precision scores of 67.9\% and 63.1\% respectively.  
\end{abstract}


\section{Introduction}


Colonial waterbird nesting islands can be found across the globe, and each of North America’s coasts is home to its own species of breeding colonial waterbirds. Colonial waterbirds are important indicators of ecosystem health \cite{burgerProductivityWaterbirdsPotentially2018}, provide numerous ecosystem services, and are an important part of a growing nature-based tourism sector of the economy \cite{NationalSurveyFishing2016}. Therefore, continuing research and monitoring of these species is critical to inform conservation decisions, encourage management of habitats for the benefit of colonial waterbirds, and to continue to gauge the surrounding ecosystem health. 

Monitoring of waterbirds at colonial nesting islands is a widespread technique used to track population trends. There are many colonial waterbird monitoring programs in the U.S. including the Texas Colonial Waterbird Survey, which is one of the longest running programs that monitors waterbirds across the entire Texas coast annually since 1976 \cite{blacklockTexasColonialWaterbird1979}. Censusing waterbirds on islands is no small task. Traditional monitoring studies of waterbirds have been conducted by traversing the colony on foot, surveying via boat, or surveying aerially using small, manned aircraft. Each of these methods has its own set of challenges and consequences. Surveying waterbirds by foot can disturb both the species of interest and the habitat occupied. Low vantage points of boat-based surveys can result in the risk of missing nests, particularly on larger and higher islands. Moreover, in certain conditions, accessing the islands by boat can be tricky due to inclement weather conditions. Manned aerial surveys
is the preferred technique by state and federal wildlife agencies, but these are expensive and require the proper conditions. In fact, aircraft crashes and boating accidents have been found to be the largest causes of mortality and injury among biologists in the field \cite{sasseJobRelatedMortalityWildlife2003}.

In recent years, unmanned aerial vehicles (UAVs), also referred to as drones, have presented themselves as a useful tool in wildlife management \cite{jonesivAssessmentSmallUnmanned2006, wattsSmallUnmannedAircraft2010}, including waterbird monitoring \cite{chabotPopulationCensusLarge2015, hanPossibilityApplyingUnmanned2017}. Drones allow researchers to remain safely on the ground while surveying areas of interest with both less cost and greater ease than traditional aerial surveys. In studies where this technology has been applied, the use of drones was found to result in more precise count estimates than traditional ground-based surveys \cite{hodgsonDronesCountWildlife2018}. Unfortunately, the expertise and time required to manually localize and classify species from hundreds, potentially thousands, of aerial images represents a major bottleneck. 

To alleviate this issue, we developed a object detection-based deep learning pipeline that utilizes convolutional neural networks (CNNs) \cite{girshickRichFeatureHierarchies2014, renFasterRCNNRealTime2015, linFocalLossDense2017} to precisely localize and classify colonial waterbird species from UAV aerial imagery via supervised learning. We collect survey images from three colonial nesting islands along the Texas coast, and train a CNN-based object detection model to detect 16 classes of waterbirds, including the 14 most common colonial waterbird species found on these islands: Brown Pelican (Pelecanus occidentalis), Laughing Gull (Leucophaeus atricilla), Royal Tern (Thalasseus maximus), Sandwich Tern (Thalasseus sandvicensis), Great Egret (Ardea alba), Cattle Egret (Bubulcus ibis), Snowy Egret (Egretta thula), Reddish Egret (Egretta rufescens), American White Ibis (Eudocimus albus), Great Blue Heron (Ardea Herodias), Black-crowned Night Heron (Nycticorax nycticorax), Tri-colored Heron (Egretta tricolor), Roseate Spoonbill (Platalea ajaja), and Black Skimmer (Rynchops niger). We present results using two of the most commonly implemented CNN-based object detection models, Faster R-CNN \cite{renFasterRCNNRealTime2015} and RetinaNet \cite{linFocalLossDense2017}. 

\subsection{Contributions}

The key contributions of this work are as follows: 
\begin{itemize}
    \item We develop a deep learning pipeline to detect waterbirds from UAV aerial imagery for precise waterbird monitoring. Our pipeline is general and can be applied to other applications requiring object detection from high-resolution aerial imagery, including other wildlife monitoring applications. Our code is available at: \url{https://github.com/RiceD2KLab/Audubon_F21}
    \item We apply our method to detect 16 classes of waterbirds from UAV aerial imagery collected from nesting islands surveyed along the Texas coast. To the best of the authors' knowledge, this is one of the largest number of species detected by a single model for UAV-based waterbird monitoring research. 
    \item We present experimental results utilizing Faster R-CNN and RetinaNet object detectors. We show that we can accurately detect 3 of the most prevalent waterbird classes in our dataset ($>70\%$ of total waterbirds): Mixed Tern Adults, Laughing Gull Adults, and Brown Pelican Adult, with an interpolated average precision ($\text{AP}_{\text{IoU}=0.5}$) score of over 90\% for Faster R-CNN and 85\% for RetinaNet. Across all waterbird classes, we achieve a mean interpolated average precision ($\text{mAP}_{\text{IoU}=0.5}$) scores of 67.9\% and 63.1\% for Faster R-CNN and RetinaNet respectively. 
\end{itemize}


\section{Related Work}

In recent years, object detection, the task of localizing one or more objects in an image with corresponding classifications, has seen immense advancements largely due to the rapid development of deep learning \cite{zhaoObjectDetectionDeep2019}. State-of-the-art object detection architectures utilizing convolutional neural networks (CNN) as a `backbone' have particularly achieved much success due to a CNN's ability to learn hierarchical image features \cite{krizhevskyImageNetClassificationDeep2012, dhillonConvolutionalNeuralNetwork2020}. 
For this work we utilize two popular CNN-based object detectors: Faster R-CNN \cite{renFasterRCNNRealTime2015} and RetinaNet \cite{linFocalLossDense2017}. Nevertheless, the proposed method is general, and is easily extensible to other object detectors. 

Consequent to the success of CNN-based object detection and wide availability of open-source code, several works have utilized these methods for wildlife monitoring with unmanned aerial vehicle (UAV) imagery. Andrew \textit{et al.} \cite{andrewVisualLocalisationIndividual2017} use a R-CNN \cite{girshickRichFeatureHierarchies2014} to detect Holstein Friesian cattle from UAV imagery, proposing both a standard still-image acquisition pipeline and an extended video monitoring pipeline. Kellenberger \textit{et al.} \cite{kellenbergerFastAnimalDetection2017} use a custom one-stage detector with an AlexNet \cite{krizhevskyImageNetClassificationDeep2012} backbone to detect large animals from UAV images captured over the Kuzikus wildlife reserve park in Namibia. Gray \textit{et al.} \cite{grayDronesConvolutionalNeural2019} use a Mask R-CNN \cite{heMaskRCNN2017} to detect and segment humpback whales, minke whales, and blue whales from UAV imagery collected off the coast of California and along the Western Antarctic Peninsula. 

More related to this work, researchers have also utilized CNN-based object detectors for bird monitoring. As compared to the aforementioned works, which focused on detecting relatively large and distinct mammals, bird detection from UAV imagery is generally regarded as a more challenging detection task due to the unique characteristics of birds. In particular, visual differences between bird species may be minor, making it difficult to distinguish between them. This difficulty is heightened for partially occluded birds, such as birds with necks tucked under their wings, as key visual features used to make distinctions are hidden.  
Borowicz \textit{et al.} \cite{borowiczMultimodalSurveyAdelie2018} use DetectNet \cite{taoDetectNetDeepNeural2016} to count Ad\'{e}lie penguins in the Danger Islands off the northern tip of the Antarctic Peninsula. Hong \textit{et al.} \cite{hongApplicationDeepLearningMethods2019} survey various off-the-shelf CNN-based object detectors to detect birds from UAV imagery collected of both wild and decoy birds in various environments across South Korea. Hayes \textit{et al.} \cite{hayesDronesDeepLearning2021} use a RetinaNet to detect seabirds, specifically Black-browed Albatrosses and Southern Rockhopper Penguins, from UAV imagery collected of the Falkland (Malvinas) Islands. 

This work expands on the existing literature of deep learning-based object detection for bird monitoring by significantly increasing the number of bird species detected by a single object detector. The aforementioned works focus on identifying 2 or fewer bird classes that are often specific to the surveyed islands. However, this work shows that CNN-based object detectors are capable of detecting several bird species, even when visually similar or imaged in challenging viewing conditions such as dense flocks or obscuring foliage. 


\section{Dataset}

\begin{figure}[t]
    \centering
    \includegraphics[width=0.9\linewidth]{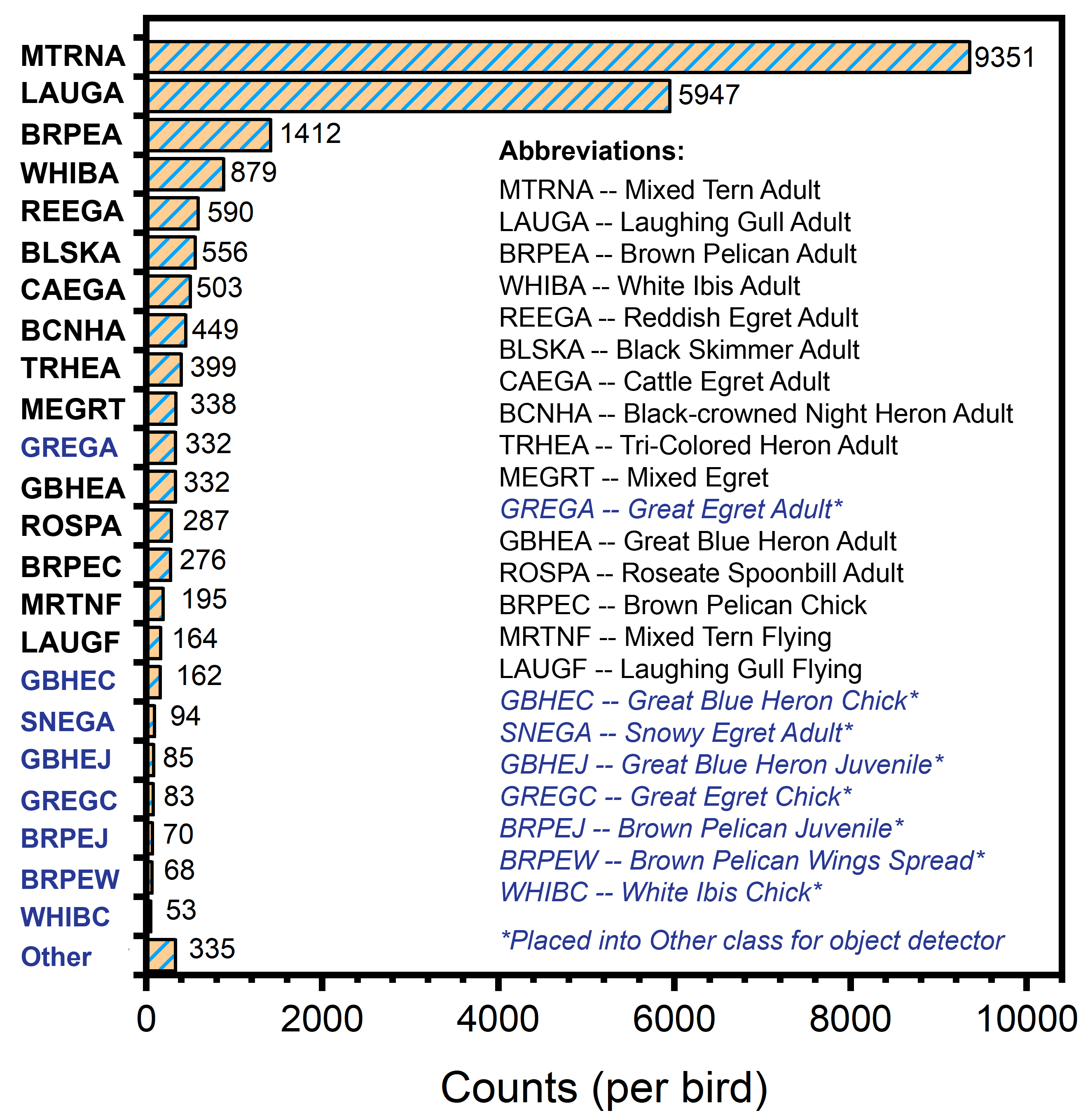}
    \caption{\textbf{Dataset distribution of waterbird classes.} 24 classes of various bird species at different ages and configurations. In this work, we select 15 unique classes to be detected by the object detector, and combine the remaining classes into the ``Other" waterbird class (highlighted by italicized, blue text).}
    \label{fig:dataset}
\end{figure}

Aerial imagery from three colonial waterbird nesting islands, Chester Island, Little Bay North Island, and North Deer Island, was captured using a DJI Matrice 300 RTK\footnote{\url{https://www.dji.com/matrice-300}} quadcopter drone with a Zenmuse P1 camera attachment\footnote{\url{https://www.dji.com/zenmuse-p1}}. A total of 200 high-resolution images are contained in the dataset, where each image is $8192 \times 5460$ pixels in resolution. Human annotations consisting of 4 bounding-box coordinates and object classes representing different waterbirds were performed for each image. Figure \ref{fig:dataset} shows the distribution of waterbird classes present in the dataset, along with a name abbreviation list. The waterbird classes were categorized based on waterbird species, maturity, and flight. Annotations were collected in this manner, as opposed to solely variations in species type, due to the large visual differences between these classes. Note that the ``Mixed Tern" and ``Mixed Egret" classes do not correspond to a single waterbird species, but rather a collection of visually similar waterbird species, for example, Royal Terns (Thalasseus maximus) and Sandwich Terns (Thalasseus sandvicensis) for the ``Mixed Tern" class. This was done due to human annotation difficulties in identifying the different species consistently in a large-scale manner. Finally, bird species that were either not of interest to the monitoring survey or that could not be identified by annotators were labelled as "Other".  

The dataset distribution is long-tailed, with the majority of waterbird classes dominated by Mixed Tern Adults and Laughing Gull Adults. From the original 24 classes, we focus detection efforts on 15 classes: Mixed Tern Adult, Laughing Gull Adult, Brown Pelican Adult, White Ibis Adult, Reddish Egret Adult, Black Skimmer Adult, Cattle Egret Adult, Black-crowned Night Heron Adult, Tri-colored Heron Adult, Mixed Egret, Great Blue Heron Adult, Roseate Spoonbill Adult, Brown Pelican Chick, Mixed Tern Flying and Laughing Gull Flying. The remaining 9 classes are all categorized as "Other". Therefore, a total of 16 classes are trained for detection by object detector.

\section{Methods}

\begin{figure}[t]
    \centering
    \includegraphics[width=0.9\linewidth]{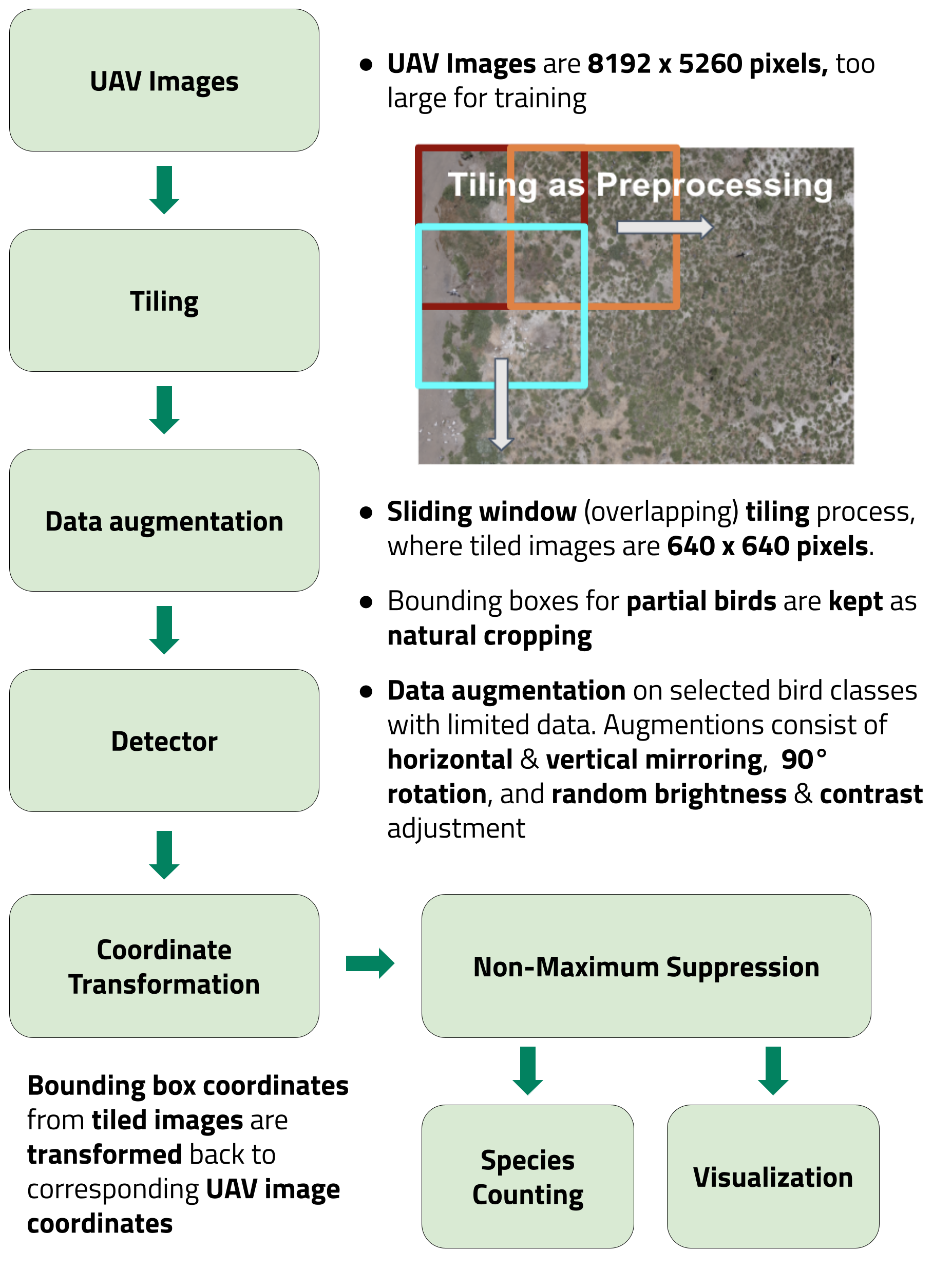}
    \caption{\textbf{Overview of the proposed pipeline for waterbird detection from UAV aerial imagery.} 
    The large raw UAV images are cropped using an overlapping sliding window tiling process. Visual data augmentations are performed to a subset of tiled images containing underrepresented waterbirds used for training the detector to alleviate class imbalance. During evaluation, bounding box coordinates predicted by the detector on the tiled images are transformed back into pixel coordinates in the large raw UAV image. Non-maximum suppression is used to filter overlapping detections in the final UAV image, enabling waterbird species counts to be obtained for each UAV image alongside visualizations. 
    }
    \label{fig:Pipeline}
\end{figure}

An overview of our proposed pipeline is shown in Figure \ref{fig:Pipeline}. Here, we discuss each component of the pipeline in detail. 

\subsection{Data pre-processing}

Given the large resolution of the original UAV images, we cropped the images using a sliding window tiling process, which is visually described in Figure \ref{fig:Pipeline}. Such cropping is necessary to ensure the object detector can be trained in a computationally feasible manner. We opted not to downsample the original UAV images to minimize information loss from the original images. Cropped image tiles are of resolution $640 \times 640$ pixels, and the sliding window shifts horizontally by 400 pixels, resulting in approximately 62.5\% overlap between adjacent tiles. Such overlap ensures at least one tile sees a complete waterbird that is also completely visible in the original UAV image. Consequently, we also keep edge tiles by adjusting the sliding window shift so the window remains within the bounds of the original image. Bounding box annotations for waterbirds cropped by the tiling process are kept if there is more than 80\% overlap in area with the original bounding box. Such partial birds are kept as a form of natural cropping as human annotators can also distinguish partially cropped or occluded birds. 

From the tiled images, we randomly split the dataset into training, validation, and testing sets following a 70-15-15\% ratio. As can be seen in Figure \ref{fig:dataset}, our dataset is highly imbalanced with 3 of the classes representing over 70\% of the total waterbird occurrences. To resolve this issue, we implement various image augmentation techniques by oversampling the images in which defined minority classes contain more than 80\% of the total cropped image annotation. The minority classes are defined to be: Brown Pelican Adults, White Ibis Adults, Reddish Egret Adults, Tri-colored Heron Adults, Great Blue Heron Adults, Roseate Spoonbill Adults, and Brown Pelican Chicks. The augmentations consist of horizontal and vertical mirroring, $90^{\circ}$ rotation, and random brightness and contrast adjustments.

\subsection{CNN-based object detector}

We explore two popular CNN-based object detection models, Faster R-CNN \cite{renFasterRCNNRealTime2015} and RetinaNet \cite{linFocalLossDense2017}, to detect waterbirds from the tiled images. A ResNet-50 feature pyramid network \cite{lin2017feature} is utilized as the backbone for both models. For implementation, we use the \texttt{Detectron2} library \cite{wu2019detectron2} for both models. We train both models using stochastic gradient descent with momentum ($\mu=0.9$) \cite{sutskever2013importance} and a batch size of 8 for a total of 1400 steps. Both models are initialized with pre-trained COCO object detection \cite{lin2014microsoft} model weights. For Faster R-CNN, the initial learning rate is $0.01$, which is then reduced by a factor of $0.01$ after 900 steps. For RetinaNet, the initial learning rate is $0.001$, which is then reduced by a factor of $0.001$ after 900 steps. The learning rate and decay factor are selected using a Bayesian hyperparameter tuning algorithm \cite{optuna_2019}. Both models use the default \texttt{Detectron2} L-2 weight regularization with regularization strength, $\lambda=1 \times 10^{-4}$. 

\subsection{Data post-processing}

When deploying the waterbird detector post-training, we transform the coordinates of the predicted bounding boxes in the tiled images back to the corresponding pixel coordinates of the original UAV image. Given the overlapping nature of the tiling process, there may be many detections for the same waterbird in the UAV image. To filter these overlapping bounding boxes, we utilize non-maximum suppression (NMS) with an intersection over union (IoU) threshold of $0.5$ \cite{malisiewicz-iccv11}. With the final predicted detections in the original UAV image coordinate space, a precise waterbird count can be made for the entire UAV image without double-counting. We note that users with access to corresponding geospatial positioning (GPS) coordinates for captured UAV images can further transform the detections to GPS coordinates. By combining detections from multiple UAV images taken during a flight mission, and filtering overlapping detections using NMS, a precise waterbird count for the area covered during the mission can be obtained.

\subsection{Evaluation metrics}

\begin{figure}[t]
    \centering
    \includegraphics[width=0.9\linewidth]{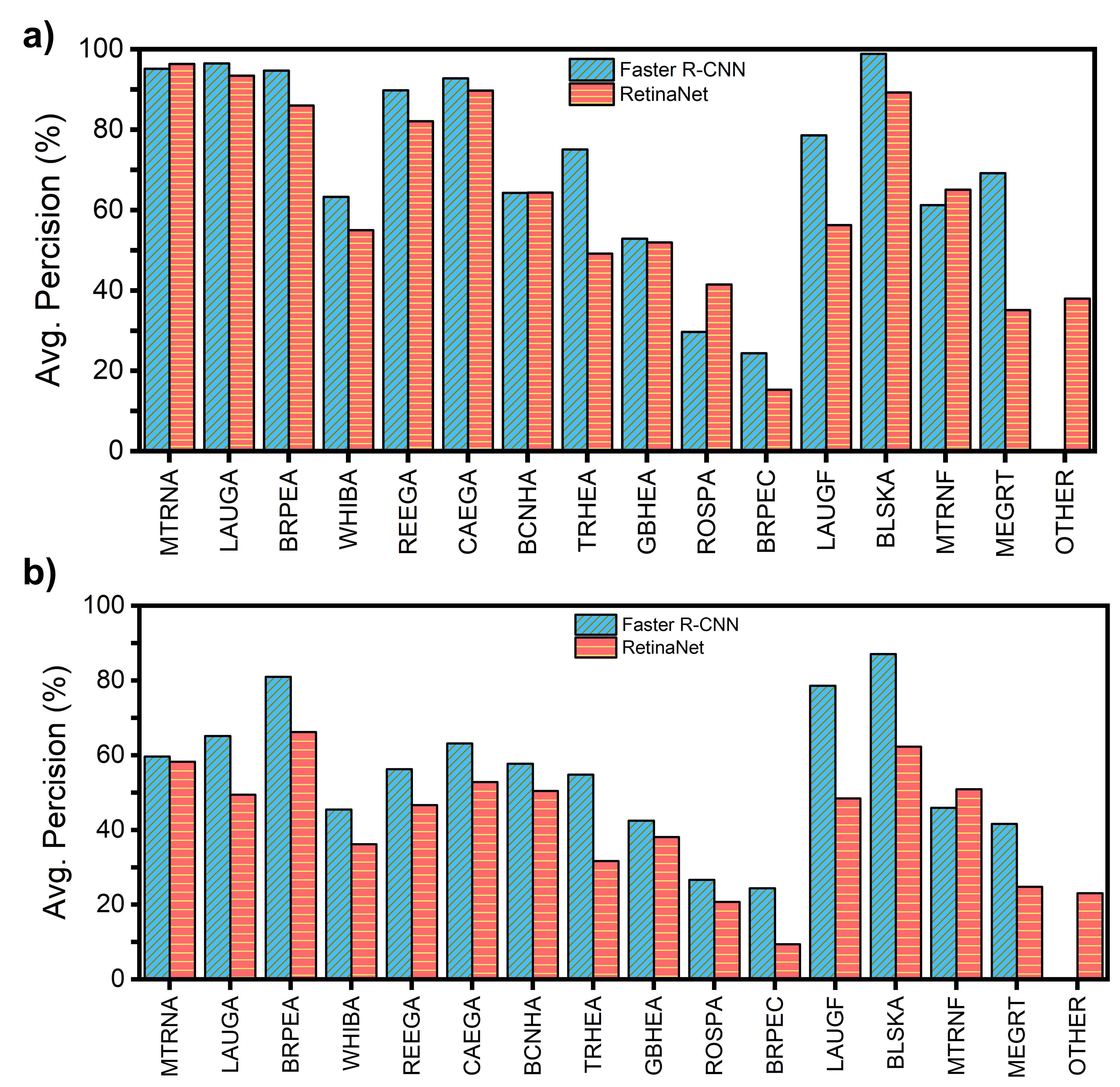}
    \caption{\textbf{Object detector interpolated AP performance for each waterbird classes.} 
    \textbf{A.} Bar chart for interpolated AP results using an IoU threshold of $0.5$. \textbf{B.} Bar chart for interpolated AP results using an IoU threshold of $0.75$.}
    \label{fig:avg_precision}
\end{figure}

To evaluate the performance of the object detectors quantitatively, we apply the interpolated average precision (AP) metric using an IoU threshold of $0.5$ and $0.75$. The interpolated AP metric is commonly used by object detection competitions such as the PASCAL VOC challenge \cite{everinghamPascalVisualObject2010} and the Microsoft COCO challenge \cite{lin2014microsoft}. The metric defines a true positive detection to be a prediction where: (i) the confidence score of the detection is greater than a threshold, (ii) the predicted class matches the ground truth class, and (iii) the IoU between the predicted and ground truth bounding box is greater than a threshold. Here, IoU is defined mathematically as: 
\begin{equation}
    \text{IoU} = \frac{\text{Area}(B_p \cap B_{gt})}{\text{Area}(B_p \cup B_{gt})}
    \label{eqn:IoU}
\end{equation}
where $B_p \cap B_{gt}$ and $B_p \cup B_{gt}$ denote the intersection and union of the predicted ($B_p$) and ground truth ($B_{gt}$) bounding boxes respectively. Precision and recall are defined as: 
\begin{equation}
    \text{Precision} = \frac{\text{True positive}}{\text{True positive} + \text{False positive}}
    \label{eqn:precision}
\end{equation}
\begin{equation}
    \text{Recall} = \frac{\text{True positive}}{\text{True positive} + \text{False negative}}
    \label{eqn:recall}
\end{equation}

From Equations \ref{eqn:IoU}, \ref{eqn:precision}, and \ref{eqn:recall}, the interpolated AP is calculated as follows: 
\begin{equation}
    \text{AP}_{\text{IoU}=t} = \sum_{r \in [0:0.01:1]} (r_{i+1} - r_{i}) p_{interp}(r_{i+1})
\end{equation}
where $t$ denotes the IoU threshold (typically ranging from $0.5:0.05:0.95$), $r$ denotes the cutoff recall, and $i$ is the step size for recall (\@ 0.01 in our case). By calculating the AP using the formula given above, we are reducing the amount of variance ``wiggle" in the precision-recall curve. The interpolated precision, $p_{interp}(r)$, is defined for the maximum recall cutoff, $r$, and is defined as the following:
\begin{equation}
    p_{interp}(r) = \max_{\Tilde{r}:\Tilde{r} \geq r} p(\Tilde{r})
\end{equation}

\begin{figure}[t]
    \centering
    \includegraphics[width=0.9\linewidth]{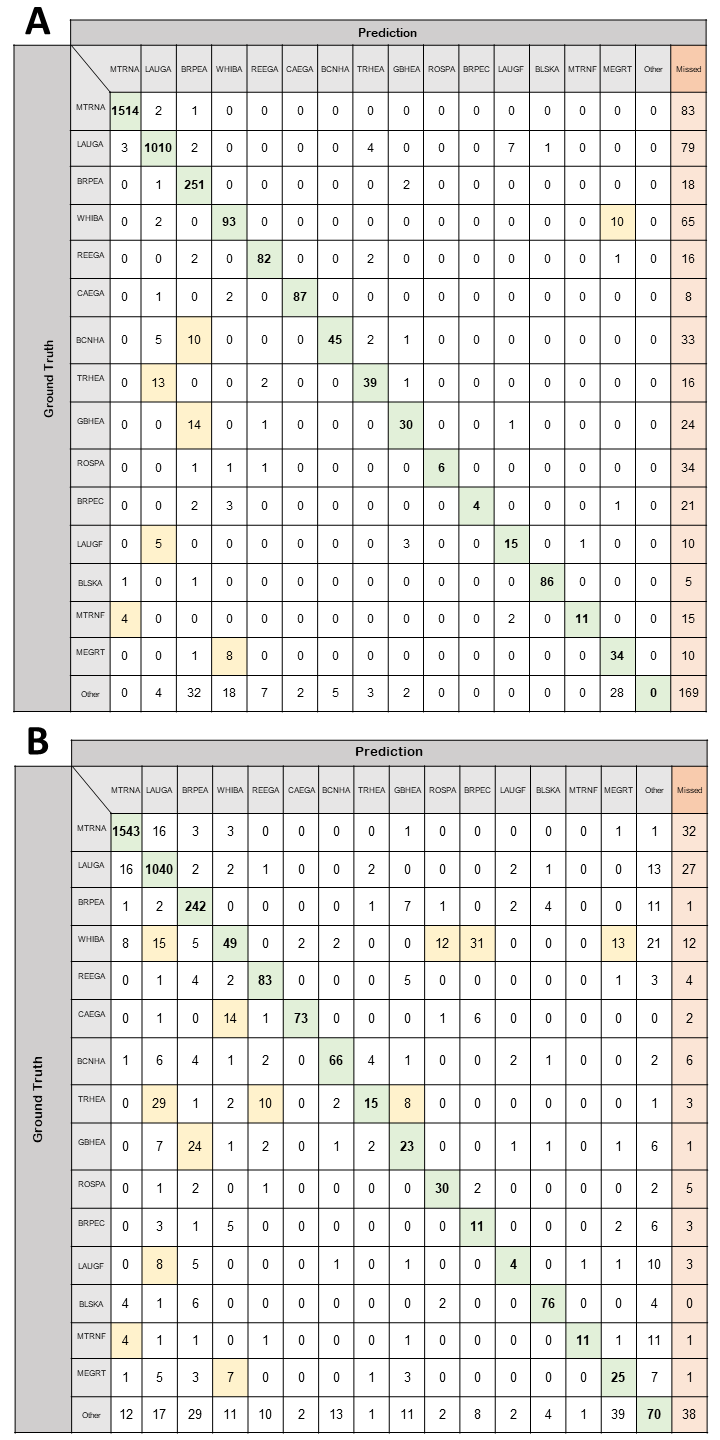}
    \caption{\textbf{Confusion matrices for object detectors}. 
    \textbf{A.} Confusion matrix for Faster R-CNN model. \textbf{B.} Confusion matrix for RetinaNet model.}
    \label{fig:confusion_matricies}
\end{figure}

In addition to interpolated AP, we construct confusion matrices to analyze the model's classification ability between different waterbird classes that the model has already correctly localized. We define a predicted waterbird detection to be a detection with a confidence score above $0.5$ and an IoU overlap between ground truth and prediction to be above $0.5$. An IoU threshold of $0.5$ is in line with PASCAL VOC metrics \cite{everingham_pascal_2015}. We include an additional column in the matrix to highlight missed detections by subtracting the total ground truth waterbirds to total number of predicted waterbirds that correspond with that species. 



\section{Experimental Results}

The interpolated AP metrics per waterbird class for both Faster R-CNN and RetinaNet are shown in Figure \ref{fig:avg_precision}, and the confusion matrices for Faster R-CNN and RetinaNet are shown Figure \ref{fig:confusion_matricies}. We find that for both models, 6 out of 16 waterbird classes are accurately detected with high interpolated AP scores ($>80$\% for IoU$=0.5$, $>60$\% for IoU$=0.75$): Brown Pelican Adults, Laughing Gull Adults, Mixed Tern Adults, Reddish Egret Adults, Cattle Egret Adults, and Black Skimmer Adults. These 6 classes correspondingly are within the top 7 classes with the highest number of occurrences in the dataset. 
With regards to which waterbirds both models detect poorly, small and often hidden waterbirds, such as Brown Pelican Chicks, Black-crowned Night Heron Adults, and Tri-colored Heron Adults, have below average interpolated AP scores. In particular, waterbird chicks are commonly found hidden behind nests, under vegetation, or under adult waterbirds, making it difficult to accurately detect such birds even by humans. 
Additionally, as expected, both models also perform poorly on waterbird classes with limited data such as the Roseate Spoonbill Adults and Brown Pelican Chicks, even after data augmentation, as these birds constitute less than 4\% of the total population. Finally, using the confusion matricies, we find that both models become confused between waterbird classes that are visually difficult to distinguish. Both models have relatively large confusion between White Ibis Adults and Mixed Egrets, both of which consist of waterbirds with white bodies of relatively similar proportions. 
Similarly, Great Blue Heron Adults are commonly misclassified by the models as Brown Pelican Adults, likely due to the similar body dimensions and relative beak size when viewed from above. Lastly, we note that there is confusion between identical waterbird species that appear in different classes due to the construction of class labels to separate waterbirds in flight -- namely, we see that the Laughing Gull Flying and Mixed Tern Flying classes are confused with the Laughing Gull Adult and Mixed Tern Adult classes respectively. 

Overall, we find that the Faster R-CNN model outperforms RetinaNet with respect to the interpolated AP metrics, which agrees with benchmark results on standard object detection datasets and competitions. However, the confusion matrices show that, in general, Faster R-CNN has a higher false negative, or missed detection, rate than RetinaNet -- specifically $14.6$\% over all waterbirds in comparison to RetinaNet's $3.3$\%. Moreover, the confusion matrix in Figure \ref{fig:confusion_matricies}A shows that Faster R-CNN fails to utilize the ``Other" waterbird class at all. Finally, we find that RetinaNet can better localize waterbirds at an IoU threshold of $0.5$ when compared to Faster R-CNN, but struggles to correctly classify the localized waterbird. 

\begin{figure*}[ht]
    \centering
    \includegraphics[width=0.8\textwidth]{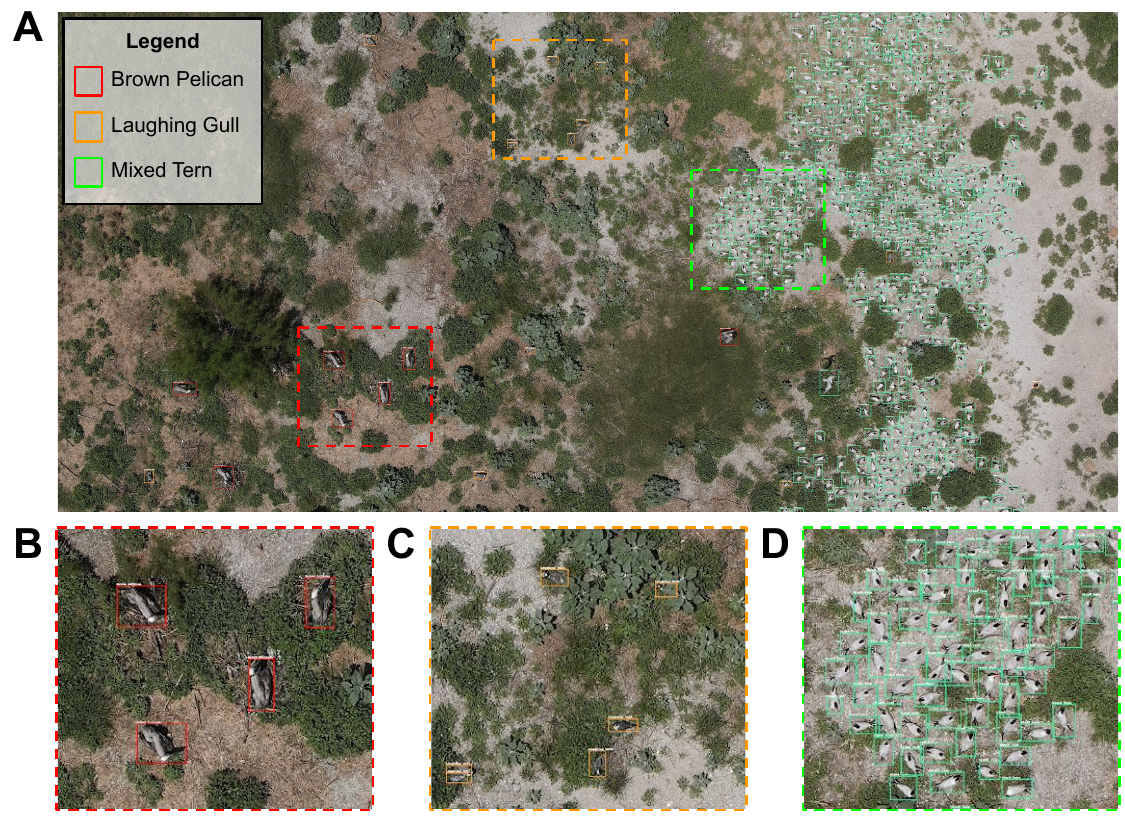}
    \caption{\textbf{Visualization of the outputs for the proposed waterbird detection from UAV aerial imagery method using Faster R-CNN.} \textbf{A.} Original high-resolution UAV aerial image with predicted waterbird detections for 3 of the most common waterbird classes present on the surveyed islands. \textbf{B.} Zoomed-in image showing predictions on a flock of Brown Pelican Adults. \textbf{C.} Zoomed-in image showing predictions on a flock of Laughing Gull Adults. \textbf{D.} Zoomed-in image showing predictions on a flock of ``Mixed Tern" Adults.}
    \label{fig:results-images}
\end{figure*}



\section{Discussion}

In this work, we develop a deep learning-based tool to monitor waterbirds from UAV aerial imagery. We apply our method for precise detection of 16 classes consisting of the most common colonial waterbird species found on nesting islands along the Texas coast. To the best of the authors' knowledge, this is one of the largest works towards combining deep learning and UAV aerial imagery for wildlife monitoring with respect to the number of species detected by a single model. We perform experiments using two popular CNN-based object detectors, Faster R-CNN \cite{renFasterRCNNRealTime2015} and RetinaNet \cite{linFocalLossDense2017}. Figure \ref{fig:results-images} visualizes sample waterbird detections on a UAV image using Faster R-CNN, illustrating the robustness at detecting waterbirds that are densely packed together or partially occluded by foliage. We quantitatively evaluate both Faster R-CNN and RetinaNet models and observe strong performance for the 3 most common waterbird classes appearing in our collected dataset: Mixed Tern Adults, Laughing Gull Adults, and Brown Pelican Adults. Specifically for Faster R-CNN, we observe this model avhieves over 90\% interpolated average precision on these 3 classes.   

For future work, the object detector's relatively poor performance on minority waterbirds present in the dataset must be solved. This is a problem characteristic to learning from long-tailed data distributions and is still an active area of research \cite{wangLearningModelTail2017, kang2020Decoupling}. Furthermore, we anticipate work to further increase the number of waterbird classes detected. In particular, we emphasize the importance of separating the ``Mixed" waterbird classes into their respective species, for example the ``Mixed Tern" class into Royal Terns (Thalasseus maximus) and Sandwich Tern (Thalasseus sandvicensis). Such a task is extremely challenging as distinct bird species within the same family, and sometimes not within the same family, appear visually similar, even causing human experts to have trouble distinguishing species apart. This problem of fine-grained recognition is also an area of active research \cite{wah2011caltech, branson2014bird, Krause_2015_CVPR}. Finally, an analysis of the trade-off between detection precision and image resolution is necessary to understand what camera specifications and flight mission altitudes are required to monitor waterbirds adequately. This is important as care must be taken in selecting flight altitudes when surveying birds with UAVs in order to minimize disturbance, follow monitoring guidelines \cite{vasApproachingBirdsDrones2015}, and ensure affordable UAV and camera equipment can be utilized by surveyors.   
 
In conclusion, we presented an artificial intelligence tool that demonstrates promising preliminary results for monitoring colonial waterbirds from UAV aerial imagery. The use of UAV aerial imagery and deep learning can advance current monitoring efforts with precise population counts in an efficient, safe, and timely manner, thereby enabling insights into more fine-scale changes in nesting habitat use year-to-year. We hope the impact of our work can help researchers and wildlife agencies alike better monitor waterbirds to gauge the ecosystem health, inform environmental policies, and protect the natural habitats of waterbirds.  

\section*{Acknowledgments}
The project is being supported by the faculty and staff of D2K Lab at Rice University. W.L. acknowledges the National Science Foundation Graduate Research Fellowship Program. This material is based upon work supported by the National Science Foundation Graduate Research Fellowship Program under grant no. NSF 20-587. Any opinions, findings and conclusions or recommendations expressed in this material are those of the author and do not necessarily reflect the views of the National Science Foundation. 

\bibliographystyle{IEEEtran}
\bibliography{references}

\end{document}